\documentclass{article}
\usepackage{spconf,amsmath,graphicx,pdfpages,enumitem}
\usepackage{hyperref,tabto}

\title{Weakly Supervised Prostate TMA Classification via \\Graph Convolutional Networks}
\name{Jingwen Wang$^1$,
        Richard J. Chen$^1$,
        Ming Y. Lu$^1$,
        Alexander Baras$^2$,
        Faisal Mahmood$^{1}$}
\address{$^1$Department of Pathology, Brigham and Women's Hospital, Harvard Medical School, Boston, MA\\
$^2$Department of Pathology, Johns Hopkins School of Medicine, Baltimore, MD\\
\url{jwang111@bwh.harvard.edu}\quad \url{faisalmahmood@bwh.harvard.edu} }
\begin{document}
\setlength{\abovecaptionskip}{-10pt}
%
\maketitle
\begin{abstract}
Histology-based grade classification is clinically important for many cancer types in stratifying patients distinct treatment groups. In prostate cancer, the Gleason score is a grading system used to measure the aggressiveness of prostate cancer from the spatial organization of cells and the distribution of glands. However, the subjective interpretation of Gleason score often suffers from large interobserver and intraobserver variability. Previous work in deep learning-based objective Gleason grading requires manual pixel-level annotation. In this work, we propose a weakly-supervised approach for grade classification in tissue micro-arrays (TMA) using graph convolutional networks (GCNs), in which we model the spatial organization of cells as a graph to better capture the proliferation and community structure of tumor cells. As node-level features in our graph representation, we learn the morphometry of each cell using a contrastive predictive coding (CPC)-based self-supervised approach. We demonstrate that on a five-fold cross validation our method can achieve $0.9659\pm0.0096$ AUC using only TMA-level labels. Our method demonstrates a 39.80\% improvement over standard GCNs with texture features and a 29.27\% improvement over GCNs with VGG19 features. Our proposed pipeline can be used to objectively stratify low and high risk cases, reducing inter- and intra-observer variability and pathologist workload. 

\end{abstract}
\section{Introduction}
\label{sec:intro}
\vspace{-3mm}
The subjective interpretation of histology slides is the standard-of-care for prostate cancer detection and prognostication. The commonly used Gleason score, which informs the aggressiveness of prostate cancer, is based on the architectural pattern of tumor tissues and the distribution of glands. Gleason grade ranges from 1 to 5 to suggest how much the prostate tissue looks like healthy tissue (lower score) versus abnormal tissue (higher score). If the glands are small, discrete and uniform, the Gleason grade is relatively lower. On the contrary, if the glands are fused or poorly-formed, the Gleason grade is higher. A primary and a secondary grade which range from 1 to 5 are assigned to prostate tissue and the sum of the two grades determines the final Gleason score. However, its manual assignment is based on visual, microscopy-based evaluation of cellular and morphological patterns, which can be extremely error-prone and time-consuming for pathologists and suffers from interobserver and intraobserver variability \cite{fuchs2011computational}. Deep learning has been widely applied to the detection of malignancies in histology images \cite{litjens2016deep, fakoor2013using, madabhushi2016image}. Researchers have demonstrated that convolutional neural networks (CNNs) can serve as a tool for automated histology image classification \cite{ghaznavi2013digital, bar2015chest, ertosun2015automated}, including Gleason score assignment, which alleviates the aforementioned limitations of subjective interpretation. However, CNNs are not sufficiently context-aware and do not capture the dense spatial relationships between cells that are predictive of cancer grade and proliferation. In addition, CNN-based methods require detailed pixel or patch level annotations which are tedious and time consuming to curate \cite{arvaniti2018automated}. 

On the contrary, graphs are a reasonable and natural option to model the distribution of the cells in histopathology images. With the cells as nodes and the edges generated by a proper algorithm, graphs can accurately capture the distribution of cells and the spatial relations between the cells. GCN is a deep learning approach for performing feature extraction and classification on graphs\cite{kipf2016semi}. In this paper, we propose a GCN based-approach for automatic patient stratification trained using TMA-level labels and node features learned via a self-supervised technique known as contrastive predictive coding (CPC) \cite{CPC,lu2019semi}. GCNs can learn from the global distribution of cell nuclei, cell morphometry and spatial features without requiring exhaustive pixel-level annotation. 
\vspace{-3mm}
\section{related work}
\vspace{-3mm}
\textbf{Objective Gleason score grading: }
Previous work on computer-aided Gleason score grading has used machine learning with pixel-level annotations. Khurd \textit{et al.}  \cite{khurd2010computer} have developed a texture classification system for automatic and reproducible Gleason grading. They use random forests to cluster extracted filter responses at each pixel into basic texture elements (textons) and characterize the texture in images belonging to a certain tumor grade. del Toro \textit{et al.} \cite{del2017convolutional} stratify the patients into high-risk and low-risk with the decision boundary of Gleason score 7-8. They train a CNN that requires pixel-level labels on regions of interests (ROI) extracted from WSIs and are able to detect prostatectomy WSIs with high–grade Gleason score. Arvaniti \textit{et al.} \cite{arvaniti2018automated} present a deep learning approach for automated Gleason grading of prostate cancer TMAs and are able to assign both the primary score and secondary score to a single pixel in the patients' TMAs. However, all current methods require pixel or patch level labels and do not incorporate the spatial organization of cells to fullt capture the architexture of the tumor micro-environment.

\noindent\textbf{Classification using GCNs: } GCNs proposed by Kipf \textit{et al.} \cite{kipf2016semi} has been utilized in many classification tasks. Recently, Zhou \textit{et al.} \cite{zhou2019cgc} propose using GCN for colorectal cancer classification, where each node is represented by a nucleus within the original image and cellular interactions are denoted as edges between these nodes according to node similarity. Then they extract spatial features like the centroid coordinates and texture features like angular second moment (ASM) obtained from grey level co-occurrence matrix (GLCM) \cite{albregtsen2008statistical, mohanaiah2013image} for each nodes. GCNs are utilized to classify each image into normal, low-grade and high-grade based on the degree of gland differentiation. Anand \textit{et al.} \cite{gadiya2019histographs} also propose to classify cancers using GCNs by modeling a tissue section as a multi-attributed spatial graph of its constituent cells. They not only extract the spatial and texture features but also use a pre-trained VGG19 model to generate features from a window of size 71 $\times$ 71 around the nuclei centroids. Then, a GCN is trained on the breast cancer BACH dataset \cite{aresta2019bach} to classify patients into cancerous or non-cancerous. However, these methods rely on gray-scale on ImageNet features and do mot use unsupervised features.
\vspace{-4mm}
\section{Methods}
\label{sec:format}
\vspace{-4mm}
Our idea is inspired by the process through which pathologists examine prostate TMAs and assign Gleason scores. The pathologists inspect the spatial distributions of the glands in TMAs. In order to model the distribution of the nuclei, we segment out the nuclei, and then construct the graphs for TMAs with nuclei as the nodes and the potential connection between neighbor nuclei as the edges. With the carefully constructed cell graphs, we are able to apply GCNs and obtain the stratification results.
\begin{figure*}[htb]
\begin{center}
  \centering
  \centerline{\includegraphics[width=\linewidth]{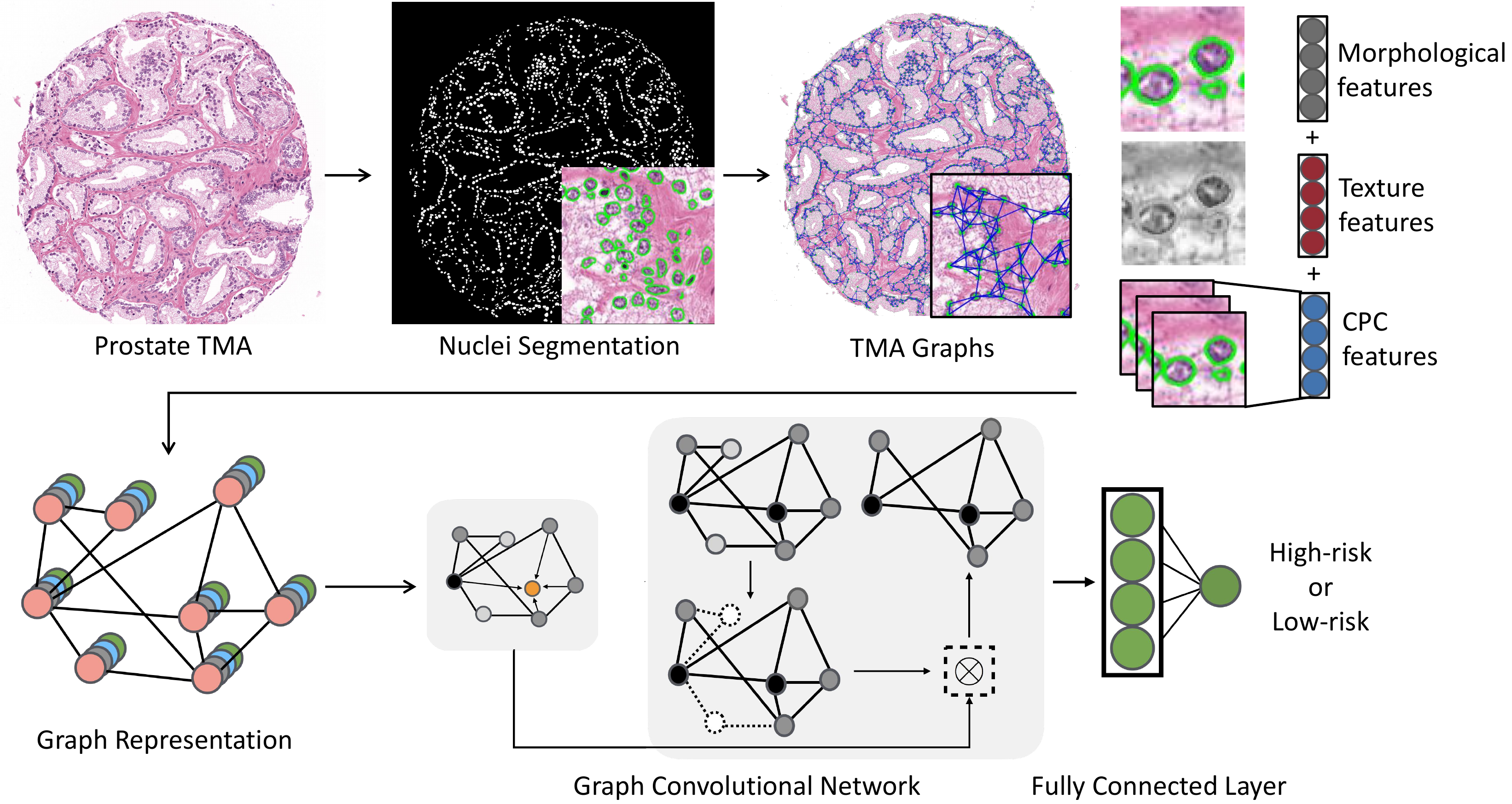}}
  \vspace{-1cm}
\end{center}
\caption{For cell identification nuclei are segmented in each TMA and a graph is built on the centeroid of each nuclei using the K-nearest neighbours (KNN) algorithm. Morphological, texture and contrastive predictive coding (CPC) features are then extracted for each nucleus and GCNs are applied on the graph representation. Finally, a fully connected layer is used at the end for classification.}
\vspace{-4mm}
\end{figure*}

\vspace{-4mm}
\subsection{TMA Graph Construction}
In order to capture the information stored in prostate TMAs, the images first need to be stain-normalized to remove the color variance. Then, each images is converted to a cell graph where the nuclei are the nodes and the possible interactions are the edges. We constructed a graph for each TMA using the following steps: i) segmenting the nuclei, which form the set of nodes. ii) modeling interactions between nuclei using edges. iii) generating features for each node.

\noindent\textbf{Nuclei Segmentation: }
To detect atypical and packed nuclei and other features that suggest cancerous tissues, precise segmentation results are fundamental. Fully convolutional neural networks (FCNN) have been used for nuclei segmentation in the past, which aims to minimize pixel-wise loss \cite{naylor2017nuclei}. However, this can sometimes leads to the fusion of nuclei boundaries and these blurry outlines of nuclei can cause feature extraction and graph structure construction to be inaccurate. In order detect the nuclei robustly and clearly, we utilize conditional GANs (cGANs) \cite{mahmood2019deep}. Such methods, which use an proper loss function for semantic segmentation, have been proved to avoid the aforementioned issues. We train our model for segmentation on the Multi-Organ Nuclei Segmentation dataset \cite{Kumar2017}. This dataset, which consists of 21,623 nuclei in 30 images, comes from 18 different hospitals and include diverse nuclear appearances from a variety of organs like liver, prostate, bladder, etc.

\noindent\textbf{Nuclei Connection: }
We assume that nuclei that are close to each other are more likely to have interactions and we intend to capture the architectural structure between neighboring nuclei. Based on this idea, we adopt the $K$-nearest neighbors (KNN) algorithm\cite{dudani1976distance}, in which each nucleus is connected to its top 5 nearest neighboring nucleus if they are within a certain Euclidean distance (100 pixels in our method). 

\noindent\textbf{Nuclei Feature Extraction: }
Based on the nuclei masks obtained using a cGAN, we then extract both morphological and texture features for each nucleus, along with features extracted from CPC-based self-supervise learning. 
\vspace{-2mm}
\begin{itemize}[leftmargin=*]
  \setlength\itemsep{0.1em}
\item \textbf{Morphological Features: }For morphological features, we compute area, roundness, eccentricity, convexity, orientation, etc for each of the nucleus.  

\item \textbf{Texture Features: }To analyze the nuclei texture in the TMAs, we calculate GLCM for each nucleus and then compute second order features like dissimilarity, homogeneity, energy and ASM based on the obtained GLCM.

\item \textbf{Contrastive Predictive Coding Features: } Contrastive predictive coding is an unsupervised learning approach to extract useful representations from high-dimensional data. By predicting the future in the latent space, CPC \cite{lu2019semi} is able to learn such representations through autoregressive models. With a certain data sequence $\{x_t\}$, a feature network $g_{enc}$ first calculates a low-dimensional embedding $z_t$ for each observation. Then, an auto-regressive context network $g_c$ can accumulate observations prior to $t$, namely, $c_t = g_c(\{z_i\}), \text{for } i \leq t$. CPC utilizes a contrastive loss, through which the mutual information shared between the context $c_t$, the present, and future observations $z_{t+k}, k>0$ can be maximized. The network aims to correctly distinguish the positive target $z_{t+k}$ from sampled negative candidates. Predictions for $z_{t+k}$ can be achieved linearly using weights $W_k$: $\hat{z}_{t+k}=W_{k} c_{t}$. If the probability score for each candidate $z_i$ is assigned through a log-bilinear model, the CPC objective is just the standard cross-entropy loss for the positive target. When applying CPC to images, small, overlapping patches need to be extracted from each image. We train a CPC encoder on extracted TMA patches of size 256 $\times$ 256 and a stride of 128. We then use the trained model to extract features from a window of size 64 $\times$ 64 around the centroid of each detected nucleus.

When applying CPC to prostate TMAs at the scale of 256 $\times$ 256 patches, we expect that the context (summarized from rows of features that are visible to the network) and unknown future observations (rows of features hidden from the network) are conditionally dependent on shared high-level information specific to the underlying pathology and the tissue site at this resolution. Examples of such high-level information might include the morphology and distinct arrangement of different cell types, the tissue texture and micro-environment, etc. By minimizing the CPC objective in the latent space, we implicitly encourage learning of such shared high-level abstractions specific to prostate tissue, which one is unlikely to effectively capture by using features extracted through naive transfer learning from ImageNet. 

\end{itemize}

\begin{table*}[htb]
\caption{A comparative analysis of prostate TMA classification using various models and features.}\label{tab:aStrangeTable}
\begin{center}
\begin{tabular}{lll}
\hline
\textbf{Model}& \textbf{Accuracy} $\uparrow$& \textbf{AUC}$\uparrow$\\
\hline
GCN + GLCM features& 0.6299 $\pm$ 0.0391& 0.6909 $\pm$ 0.0240\\
\hline
GCN + GLCM + Transfer Learning (ResNet56)& 0.6412 $\pm$ 0.0181& 0.6987 $\pm$ 0.0294\\
\hline
GCN + GLCM + Transfer Learning (VGG19) \cite{gadiya2019histographs}& 0.7194 $\pm$ 0.1192& 0.7486 $\pm$ 0.0377\\
\hline
\textbf{GCN + GLCM + CPC features (Proposed)}& \textbf{0.8995 $\pm$ 0.0222}& \textbf{0.9659 $\pm$ 0.0096}\\
\hline
\end{tabular}
\end{center}
\vspace{-5mm}
\end{table*}

For each nucleus in a prostate TMA, we generate 8 morphological features, 4 texture features from GLCMs and 1024 features from CPC. Then, we concatenate them together to form a feature matrix $V \in R^{N_i\times F}$ where $N_i$ is the number of nuclei in the graph and \textit{F} is the number of features (1036 in our method). 
\vspace{-5mm}
\subsection{Graph Convolution Networks}
\vspace{-3mm}
While CNNs remain a powerful deep learning tool for medical image analysis, they can only perform on grid-like Euclidean data like 2-dimensional images. Studies have been exploring how to generalize convolutional networks onto non-Euclidean domains. Convolution operators have been refined by Kipf \textit{et al.} \cite{kipf2016semi} to perform on non-grid-like data as graphs like protein structure and nuclei distribution. Through message passing, each node is able to iteratively accumulate feature vectors from its neighboring nodes and generate a new feature vector at the next hidden layer of the network, thus GCNs can learn to represent for each feature in a node. This process is very similar to feed-forward networks or CNNs. Then, by pooling over all the nodes, we are able to achieve the presentation for the entire graph. Such graph presentation can then serve as inputs for tasks as graph-level classification.
 
In our method, we used the GraphSAGE convolution \cite{hamilton2017inductive}. The convolution and pooling operations can be defined as follows:
\vspace{-3mm}
\begin{equation*}
\begin{aligned}
    a_{v}^{(k)} &= \textbf{MAX}\left(\left\{\textbf{ReLU}\left(W \cdot h_{u}^{(k-1)}\right), \forall u \in \mathcal{N}(v)\right\}\right) \\
    h_{v}^{(k)} &= W \cdot\left[h_{v}^{(k-1)}, a_{v}^{(k)}\right]
\end{aligned}
\end{equation*}

At the $(k-1)^{th}$ iteration of the neighborhood accumulation, the feature vector for node $v$ is denoted by $h_v^{(k)}$ while the feature vector of node $v$ at the next iteration is represented by $a_v^{(k)}$.

For the pooling part, we utilize the self-attention graph pooling method \cite{lee2019self}. Self-attention can reduce computational complexity and take the topology into account. By performing self-attention graph pooling, our method is able to calculate attention scores, focus on the more meaningful node features and aggregate information of nuclei graph topology on different levels. The attention score $Z \in \mathcal{R}^{N \times 1}$ for nodes in $G$ can be calculated as follows:

\begin{equation*}
\begin{aligned}
    Z=\sigma\left(\textbf{SAGEConv}\left(X, A+A^{2}\right)\right)
\end{aligned}
\end{equation*}
The node features are represented by $X$, the adjacency matrix is suggested by $A$, while SAGEConv is the convolution operator from GraphSAGE. 
\vspace{-5mm}
\section{Experiments and Results} 
\vspace{-3mm}
We train both the CPC and the GCN on a high-resolution H\&E stained image dataset from 5 prostate TMAs, each containing 200-300 spots \cite{zhong2017curated, DVN/OCYCMP_2018}. There are 886 images in total. For each image, there is a pixel-level annotation mask suggesting the Gleason scores. In order to obtain the TMA-level labels, we calculate the primary grade and secondary grade based on the area in the mask and sum up the two grade together as the final Gleason score. For data augmentation, we took 1550 $\times$ 1550 crops from the four corners and the center of each image, and computed their corresponding binary labels. The crops that do not contain tissue are discarded. We classify the TMAs with Gleason score higher than (including) 6 as high-risk while TMAs with Gleason score lower than 6 as low-risk. Gleason scores of $3+3$ and $3+4$ are known to have the most interobserver and intraobserver variability \cite{ozkan2016interobserver} and our proposed method can stratify between the two highly variable scores. We validate our approach on 5 class-balanced splits where each splits contain 3498 crops for training, 388 crops for validation and 432 crops for testing. Table 1. shows a comparative analysis of our proposed method with other methods in a five-fold cross validation.

\begin{figure}[htb]
\begin{minipage}[a]{1\linewidth}
  \centering
  \centerline{\includegraphics[width=\linewidth]{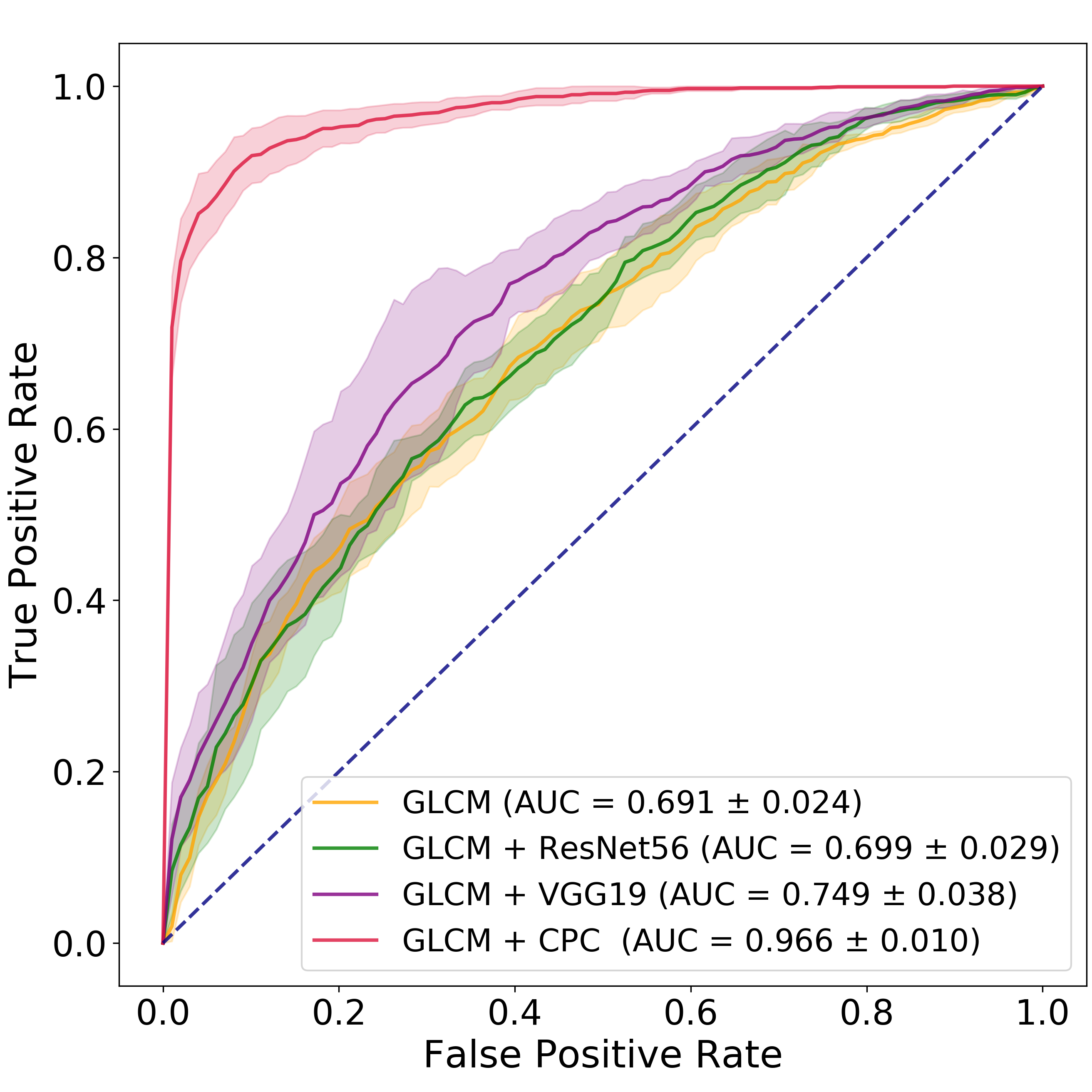}}
\end{minipage}
\vspace{-4mm}
\caption{Comparative analysis of our approach (GCN with GLCM+CPC features) as compared to other approaches demonstrates that our method achieves a 39.80\% higher AUC over GCNs with just GLCM features and a 29.27\% improvement over GCNs with VGG19 features. The shade suggests the confidence interval from the 5 random cross validation}
\vspace{-5mm}
\end{figure}

For comparison, we trained GCNs with the features generated from GLCMs, ResNet56 and VGG19 in contrast to the features generated from CPC. Results show GCNs with a combination of GLCM and CPC features achieve state-of-the-art results with an AUC of 0.9659.
\label{sec:print}
\vspace{-5mm}
\section{Conclusions}
\vspace{-5mm}
\label{sec:typestyle}
In the paper, we propose an approach for using GCNs as a tool to stratify patients on prostate cancer. Our approach is able to accurately identify high-risk patients using the prostate TMAs with only the image-level labels instead of the pixel-level labels. Also, we demonstrate that using CPC features in graph convolution can outperform features generated through simple transfer learning from ImageNet. 

Our deep learning-based TMA stratification pipeline can be used as an assistive tool for pathologists to automatically stratify out high-risk cases before review. This work has the potential to alleviate the overall pathologist burden, accelerate clinical workflows and reduce costs. Future work will involve scaling this to a six class classification problem, identifying nodes being activated for interpretability and fusing information from patient and familial histories and incorporating multi-omics information for improved diagnostic and prognostic determinations. We will also explore methods to scale this to whole slide images.



\small\bibliographystyle{IEEEbib}
\small\bibliography{refs}

\end{document}